\documentclass[letterpaper, 10 pt, conference]{ieeeconf}  % Comment this line out if you need a4paper
\IEEEoverridecommandlockouts                              % This command is only needed if 
\usepackage{amsfonts}
\usepackage{graphicx}
\usepackage{cite}
\usepackage{multirow}
\usepackage{multicol}
\usepackage{booktabs}
\usepackage{makecell}

\makeatletter
\let\NAT@parse\undefined
\makeatother
\usepackage{hyperref}  %hyperref still needs to be put at the end!

\title{\LARGE \bf DGMem: Learning Visual Navigation Policy without Any Labels by Dynamic Graph Memory}
\author{Wenzhe Cai,
        Teng Wang,
        Guangran Cheng,
        Lele Xu,
        Changyin Sun$^{\dagger}$ ,
\thanks{$\dagger$ Corresponding Author}% <-this % stops a space
\thanks{Wenzhe Cai and Guangran Cheng are with the School of Automation, Southeast University, Nanjing, China. The work is done while both are interns of PengCheng Lab.}
\thanks{Teng Wang, Lele Xu and Changyin Sun are with the School of Automation, Southeast University, Nanjing, China. (e-mail: wangteng@seu.edu.cn, xulele@seu.edu.cn, cysun@seu.edu.cn)}}% <-this % stops a space
\begin{document}
\maketitle
\thispagestyle{empty}
\pagestyle{empty}

%%%%%%%%%%%%%%%%%%%%%%%%%%%%%%%%%%%%%%%%%%%%%%%%%%%%%%%%%%%%%%%%%%%%%%%%%%%%%%%%
\begin{abstract}
    In recent years, learning-based approaches have demonstrated significant promise in addressing intricate navigation tasks. 
    Traditional methods for training deep neural network navigation policies rely on meticulously designed reward functions or extensive teleoperation datasets as navigation demonstrations. 
    However, the former is often confined to simulated environments, and the latter demands substantial human labor, making it a time-consuming process. 
    Our vision is for robots to autonomously learn navigation skills and adapt their behaviors to environmental changes without any human intervention.
    In this work, we discuss the self-supervised navigation problem and present Dynamic Graph Memory (DGMem), which facilitates training only with on-board observations.
    With the help of DGMem, agents can actively explore their surroundings, autonomously acquiring a comprehensive navigation policy in a data-efficient manner without external feedback. 
    Our method is evaluated in photorealistic 3D indoor scenes, and empirical studies demonstrate the effectiveness of DGMem.
\end{abstract}

\section{Introduction}
Navigation, a fundamental skill for mobile robots, serves as the cornerstone for basic locomotion before engaging in high-level tasks within a home-assistance robot. 
Classical methods for embodied agent navigation hinge on simultaneous localization and mapping (SLAM)\cite{LaValle2006PlanningA,Thrun2002ProbabilisticR,Wrobel2001MultipleVG}, relying on intricate hand-crafted modules and statistical analysis for robust performance.
However, the traditional lidar SLAM systems are inappropriate for abstract visual navigation tasks, such as finding the green apple on the table. 
Visual SLAM systems, dependent on accurate state estimation, are susceptible to dynamic elements, leading to mapping inconsistency.
In response to these challenges, learning-based methods offer a fresh paradigm for navigation tasks through concise end-to-end networks. 
This shift has given rise to advanced methods that have achieved notable success across various navigation tasks, including 
PointGoal Navigation~\cite{Wijmans2020DDPPOLN,Datta2020IntegratingEL}, 
ObjectGoal Navigation~\cite{du2021vtnet,Pal2020LearningHR}, 
Room Navigation~\cite{Wu2019BayesianRM,Narasimhan2020SeeingTU}, 
ImageGoal Navigation~\cite{Zhu2016TargetdrivenVN,Choi2021ImageGoalNV}, 
Vision-Language Navigation~\cite{Wang2021StructuredSM},
and Audio-Visual Navigation~\cite{Chen2020LearningTS}.

\begin{figure}[ht]
    \centering
    \includegraphics[width=0.5\textwidth]{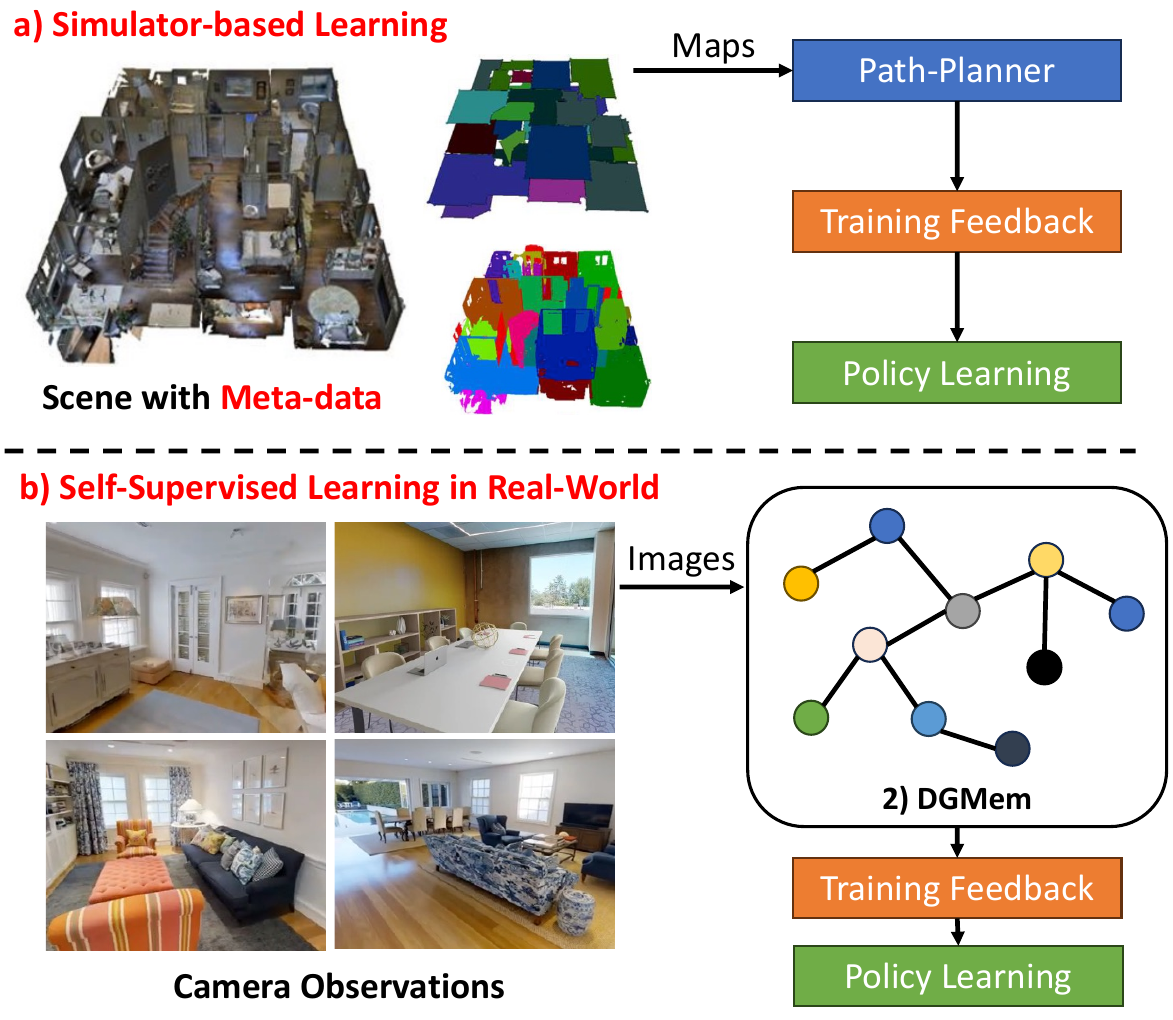}
    \caption{Explaination of the self-supervised navigation task. a) When training in the simulator, the training feedbacks come from the available meta-data. 
    But in the real-world, it is non-trivial to acquire such meta-data without human guidance for the computation of training objectives. Therefore, in b), the only accessible information is on-board sensor data, like the RGB images from the camera.
    Our DGMem must evaluate the executed actions to optimize the policy network and constantly perform navigation to collect experience.}
    \label{Fig-GraphExploration}
  \end{figure}

To learn a navigation policy, most works resort to imitation learning (IL) or reinforcement learning (RL).
Typically, training feedback is derived from human teleoperation~\cite{Ramrakhya2022HabitatWebLE} or meta-data, such as occupancy maps or shortest-path planners, embedded in simulators~\cite{Kolve2017AI2THORAI,Savva2019HabitatAP,Szot2021Habitat2T,Shen2020iGibson1A,Li2021iGibson2O}.
However, deploying a frozen pre-trained navigation policy on a real robot operating in a new environment may not yield optimal performance.
Variations in layouts and object placements necessitate the robot's ability to dynamically adjust its parameters.
Accessing meta-data used for RL in real-world scenes is impractical, and continually collecting new demonstrations using human effort to guide the robot's learning process is cumbersome.
The ideal scenario involves the robot autonomously learning through active exploration and self-supervised learning. 
In this work, we delve into the self-supervised visual navigation task, wherein the agent must acquire a generalizable navigation policy without external feedback, relying solely on its own on-board observations.

The self-supervised navigation task presents two primary challenges: first, learning the navigation policy in a data-efficient manner in the absence of external feedback; 
and second, preventing the robot from traversing a limited region to enhance the generalization performance of the learned navigation policy.
To address these challenges, we introduce a memorizing scheme, the \textbf{D}ynamic \textbf{G}raph \textbf{Mem}ory \textbf{(DGMem)}. 
DGMem dynamically iterates its stored nodes and edges as the robot roams within the house, refining itself and providing evolving feedback for both RL and IL training. 
The combination of RL+IL training objectives significantly enhances data efficiency. 
To improve generalization performance, the topological structure of DGMem reveals exploration frontiers within the scene, fostering consistent exploration behavior. 
This ensures the training dataset covers nearly all areas in the house. 
Additionally, the graph serves as a planner, aiding in the execution of long-horizon navigation trajectories and alleviating the requirements for a learning-based navigation policy. 
Both aspects are crucial for enhancing navigation performance. In summary, our contribution encompasses three folds:
\begin{itemize}
    \item We introduce the self-supervised navigation task, a crucial skill for deploying robots in the real world. 
    \item We propose DGMem, a novel memorizing scheme that serves as both a planner and trainer for the navigation policy.
    \item We evaluate DGMem in a photorealistic indoor simulator, demonstrating that even with random weight initialization, the policy network can acquire a generalizable navigation skill in the novel scene within 250k interactions.
\end{itemize}

\section{Related Works}
\subsection{Learning-Based Navigation}
With the success of end-to-end deep reinforcement learning for video games~\cite{Mnih2015HumanlevelCT},
many researchers start discussing learning-based approaches for visual navigation problems and many advanced approaches are proposed.
For example,~\cite{jaderberg2017reinforcement} accelerates navigation policy training by incorporating auxiliary learning objectives.
For a better understanding of the scene,~\cite{Fang2019SceneMT} introduces a scene memory transformer module for several navigation tasks.
~\cite{chen2018learning} utilize a mapping module and design a novel intrinsic reward for fast exploration of new scenes.
To reduce the dependency on simulators,~\cite{NEURIPS2020_2cd4e8a2,hahn2021no} use massive internet video clips as training data.
And some recent works directly using the teleoperation navigation demonstrations to train a general navigation skill~\cite{}.
A versatile navigation policy that can directly transfer to all kinds of scenes is appealing but challenging. 
For personal usage, we seldom move the robot to other houses but only expect it 'overfitting' in our own house.
Therefore, we investigate the problem of active learning without any external feedback or human intervation and we propose the DGMem which enables the robot learning navigation policy 
by itself. 

\subsection{Self-Supervised Reinforcement Learning}
The objective of reinforcement learning is maximizing the cumulative reward by adjusting a learnable policy.
Many RL-based applications are limited in simulators because only in simulators we can monitor the system status and the environment to design an appropriate reward function to guide the agent.
In most real-world scenarios, we cannot access the underlying environment states thus a well-shaped reward function is unavailable.
To tackle this problem, many works discuss a self-supervised setting for the Markov Decision Process (MDP), 
where the environment no longer gives the reward feedback, but only transits to the next state.
Under such circumstances, one possible solution is to design a general model to produce pseudo-rewards with respect to agent observations.
For example, \cite{Chen2021LearningGR} use internet video clips and a self-supervised training objective to train a discriminator as the reward generator.
\cite{touati2021learning} introduces forward-backward representation and utilizes this representation to estimate the reward.
\cite{Schwarzer2021PretrainingRF} trains a state representation with self-prediction and proposes measuring the feature similarity between state and goals as reward.
Based on mutual information, DIAYN\cite{eysenbach2018diversity,Sharma2020Dynamics-Aware} propose an intrinsic reward and learns diverse skills in an unsupervised manner.
Our work follows the reward-free settings and trains a goal-conditional navigation policy. 
But different from the prior work, our proposed DGMem can composes both IL and RL training objectives which greatly increase the data efficiency.

\subsection{External Memory for Navigation}
To complete long-horizon navigation tasks, the structural spatial information about the environment is usually important for path planning. 
Many works store key information in different types of external memory modules and exploit them for inference.
Inspired by traditional SLAM, several approaches maintain a global occupancy map as interacting with the environment.
For example, Neural SLAM\cite{Chaplot2020Learning} plans a subgoal in the global map which serves as an exploration direction.
\cite{georgakis2022learning} follows the idea and introduces semantic information to the global map. 
Some other works build the memory modules as an abstract topological map.
For example, BGM\cite{Wu2019BayesianRM} builds the topological relationship for different types of rooms.
VGM\cite{Kwon2021VisualGM} constructs the map with image observations and extract the topological-level knowledge with a neural network.
Most of the prior works only focus on improving the policy generalization capability with the external memory as policy inputs. 
But they must rely on a pre-training phase with a simulator.
But our scheme shows that DGMem can enables an agent learning the navigation policy from stretch or continue learning in a novel environment, which is unlikely to suffer from the sim-to-real domain gap.

\begin{figure*}[h!]
    \centering
    \includegraphics[width=0.95\textwidth]{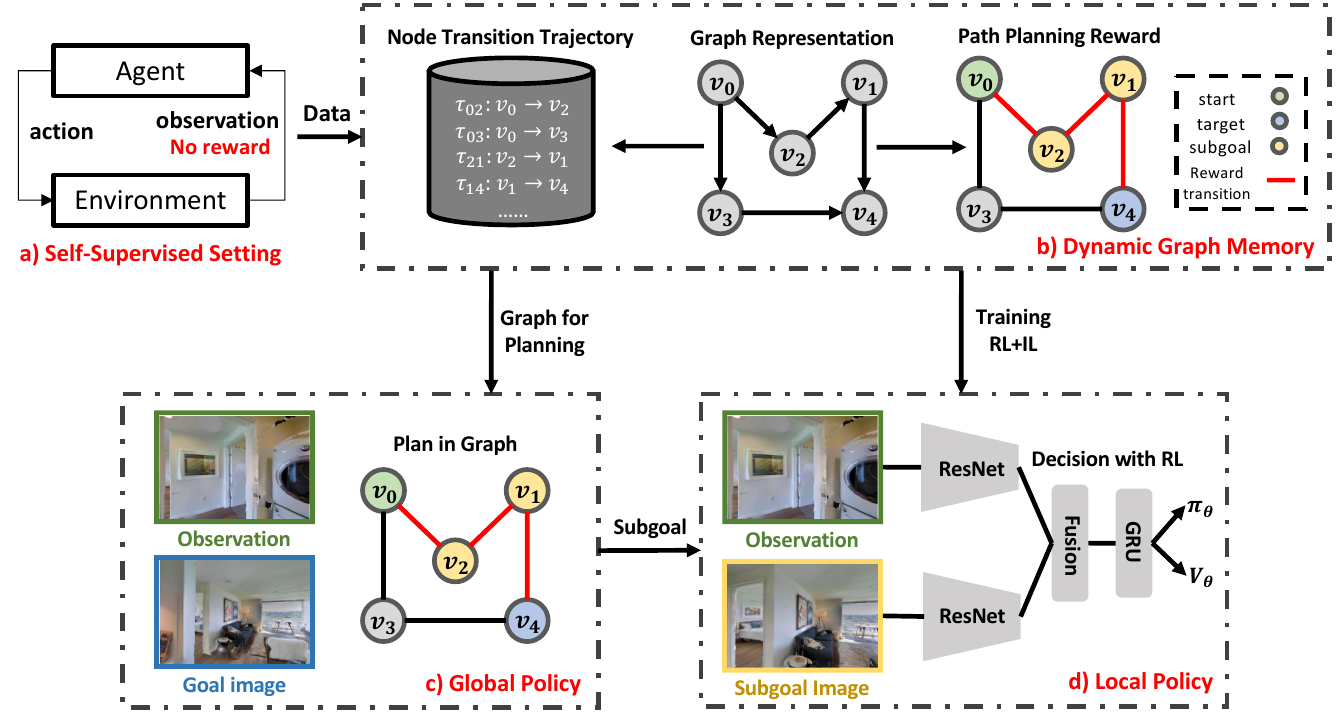}
    \caption{Framework of our approach.  a) Different from traditional RL, in a self-supervised setting, no reward function is available from the environment.
    b) As the agent interacts with the environment, it builds a graph memory from the experience data. Graph memory is used for both imitation learning and reinforcement learning.
    c) By localizing the current observation and the goal observation in the DGM, the global policy can decompose the task by planning.  Then, the next subgoal node will be conveyed to the local policy.
    d) The low-level controller will output the actions based on a trained goal-conditional policy.}
    \label{Fig-DGM}
\end{figure*}

\section{Approach}
\subsection{Problem Formulation}
We formulate the self-supervised navigation task as a Markov Decision Process (MDP).
During the training stage, we assume the agent is randomly initialized in the environment and experiencing an infinite episode, mirroring the scenario of acquiring a new home-assistant robot and deploying it in the house.
Reset functions are unavailable, simulating the constraint that the robot cannot teleport. 
The agent maneuvers through the environment by executing locomotion actions $a_{t} \in \mathcal{A}$ according to a policy $\pi_{\theta}$.
The action set $\mathcal{A}$ contains three primitive actions which are $\mathcal{A}=\{MoveAhead,TurnLeft,TurnRight\}$.
After each action $a_{t}$, the agent receives a new observation $o_{t+1}$, forming the experience trajectory $D^{train}=\{o_{0},a_{0},o_{1},a_{1}... o_{N},a_{N}\}$.
Each $o_{t}$ comprises an RGB image observation $I_{t}\in \mathbb{R}^{C\times H\times W}$ and a noisy pose estimation $\hat{P}_{t} = (\Delta x_{t}, \Delta y_{t}, \Delta z_{t}, \Delta r_{t})$,
representing relative distance along the x-y-z axis and yaw angle to the initial state. The noisy pose estimation accounts for odometry errors in real-world scenarios.
During the testing stage, the agent can be initialized at anywhere in this house. The robot must first locate itself and then perform the navigation skill.
In this work, we focus on the image-goal navigation task, where the robot is asked to navigate towards a goal image $g_{i} \in \mathbb{R}^{C\times H\times W}$.
The $g_{i}$ can be arbitrary images captured in this house, therefore, this task evaluates the generalization ability of a goal-conditioned policy $\pi_{\theta}(a_{t}|o_{t},g_{i})$.
Given the absence of external feedback, like reward function or demonstration trajectories, this poses a challenging self-supervised learning problem.
We consider training $\pi_{\theta}(a_{t}|o_{t},g_{i})$ from stretch, accessing the function of our proposed trainer DGMem, though pre-trained weights can enhance efficiency.

\subsection{Dynamic Graph Memory}
We build the DGM as an undirected graph with self-collected experience from the new environment.
The graph $\mathcal{G}$ contains a node set $\mathcal{V}$ and an edge set $E$.
In the next following sections, we will introduce the details of graph iteration, including the node updates and edge updates, respectively.

\textbf{Node Iteration.} In DGMem, each node represents an instance of the collected observations, comprising four components $v_{i}=(I_{i},\hat{P}_{i},c_{i},\Phi_{\theta}(I_{i}))$.
Here, $I_{i}$ represents the RGB image, $\hat{P}_{i}$ represents the noisy pose estimation, $c_{i}$ represents the visit count of node $v_{i}$, and $\Phi_{\theta}(I_{i})$ represents the visual encoding of the RGB image.
For efficiency, we employ a pre-trained ResNet18~\cite{He2015DeepRL} as the visual encoder $\Phi_{\theta}$.
To streamline the planning algorithm's efficiency on the graph, we limit the graph's complexity by preserving only valuable observations as nodes. 
The value of an observation is evaluated from two perspectives.
First, the semantic information conveyed in the image is considered.
We favor images with more apparent objects, utilizing a pre-trained YOLO-v5~\cite{ultralytics2021yolov5} to detect objects in the image and label the semantic score as $C^{o}_{i}=\sum_{id=0}^{K} c_{id}$, 
where $c_{id}$ represents the confidence score of the detected objects. 
Images with $C^{o}_{i} < d_{c}$ are discarded for graph iteration, where $d_{c}$ is a scalar threshold.
This ensures that background images, like those in front of a white wall, don't confuse the agent for self-localization.

Second, we consider the similarity between observations and existing nodes to decrease graph redundancy. 
We calculate both visual and pose similarity scores for each observation.
The pose similarity score $C^{e}_{i}$ is the eculidean distance between the nearest node in $\mathbb{V}$ to the observation $o_{i}$, defined as follows:
\begin{equation}
    C^{e}_{i} = \min_{\forall j \in 0,1...N} || P_{i} - P_{j} ||_{2}
\end{equation}
However, the Euclidean distance doesn’t account for the house structure. For example, two adjacent bedrooms may be close in eculidean distance, but they are two distinguished instances where we need both nodes for them.
To address this, we introduce the visual similarity score $C^{s}_{i}$, defined as the negative cosine similarity to the most similar node's visual feature.
Observations violating the rule $ C^{e}_{i} + \alpha \cdot C^{s}_{i} < d_{p} $ are discarded. 
We append the collected observation as new nodes only if both semantic and similarity scores exceed a threshold, ensuring that the graph is sparse and representative.

\textbf{Edge Iteration.} The edges in the graph represent the transition possibility from one node to another. 
We employ a binary flag to indicate reachability and dynamically update the edges based on the collected observations.
Therefore, the edge $e_{ij}$ exists only if the agent believes it is easy to traverse from the node $v_{i}$ to $v_{j}$ without bypassing any intermediate nodes.
In addition to storing the connected node indexes of the edges, we attach optimal trajectory $\tau_{ij}$ and visit count $c_{ij}$ as additional attributes of the edge.
Thus, every edge $e_{ij}$ is denoted as $e_{ij} = (i,j,\tau_{ij})$. 
To iterate the edges, at each timestep, the agent will first locate itself on the graph based on visual and pose similarity.
Denote the last localized node as $v_{t-1}$, if $ \min_{\forall j \in 0,1...N} C^{e}_{i} + \alpha \cdot C^{s}_{i} < d_{locate}$, the location will be updated to be $v_{j}$.
Otherwise, the location will keep fixed as $v_{t-1}$. The visit count $c_{ij}$ reflects the reachability between the node $v_{i}$ and $v_{j}$. 
A higher value of $c_{ij}$ means the necessity of edge $e_{ij}$ in connecting different parts of the scene. And we periodically delete those redundant edges with less $c_{ij}$ values.
As the robot working in the house, it may encounter multiple different trajectories leading from node $v_{i}$ to $v_{j}$. 
Some of these trajectories are more efficient than others, characterized by smaller transition steps.
To construct the dataset for imitation learning, we optimize $\tau_{ij}$ by retaining the most efficient trajectory while discarding others.

\subsection{Active Exploration}
To acquire a goal-conditional policy $\pi_{\theta}(a_{t}|o_{t},g_{i})$, the diversity of $o_{t}$ and $g_{i}$ in the training set plays a crucial role in determining the generalization performance of the navigation policy.
In contrast to conventional settings where dataset diversity can be ensured through explicit dataset curation or simulation, our self-supervised navigation task presents a unique challenge. 
The task cannot anticipate unseen states, necessitating active exploration to enhance dataset diversity. 
Consequently, we encourage the agent to navigate towards novel nodes, fostering exploration of new areas.

Leveraging DGMem as an abstraction of the partially observed scene, the novelty of nodes is easily defined using a count-based method.
A lower value of the node visit count $c_{i}$ signifies higher novelty.
During the training stage, the agent is periodically assigned a node as the navigation goal based on a softmax probability.
The probability $p_{i}$ of node $i$ to be chosen as the goal is defined as follows:
\begin{equation}
    p_{i} = \frac{e^{-\gamma c_{i}}}{\sum_{j=0}^{N}e^{-\gamma c_{j}}}
\end{equation}
where $\gamma$ is a temperature scalar and $N$ represents the number of the existing nodes.
The novelty of nodes is associated with the uncertainty of local areas, and navigating to these uncertain areas provides a better chance to enhance local topology and sequentially unlock new areas.
With such an active exploration scheme, the graph memory gradually covers all areas in the house, providing more comprehensive candidates for goal-conditional policy learning. 
An illustration of the exploration procedure are shown in Fig \ref{Fig-GraphExploration}.

\begin{figure}[h!]
  \centering
  \includegraphics[width=0.48\textwidth]{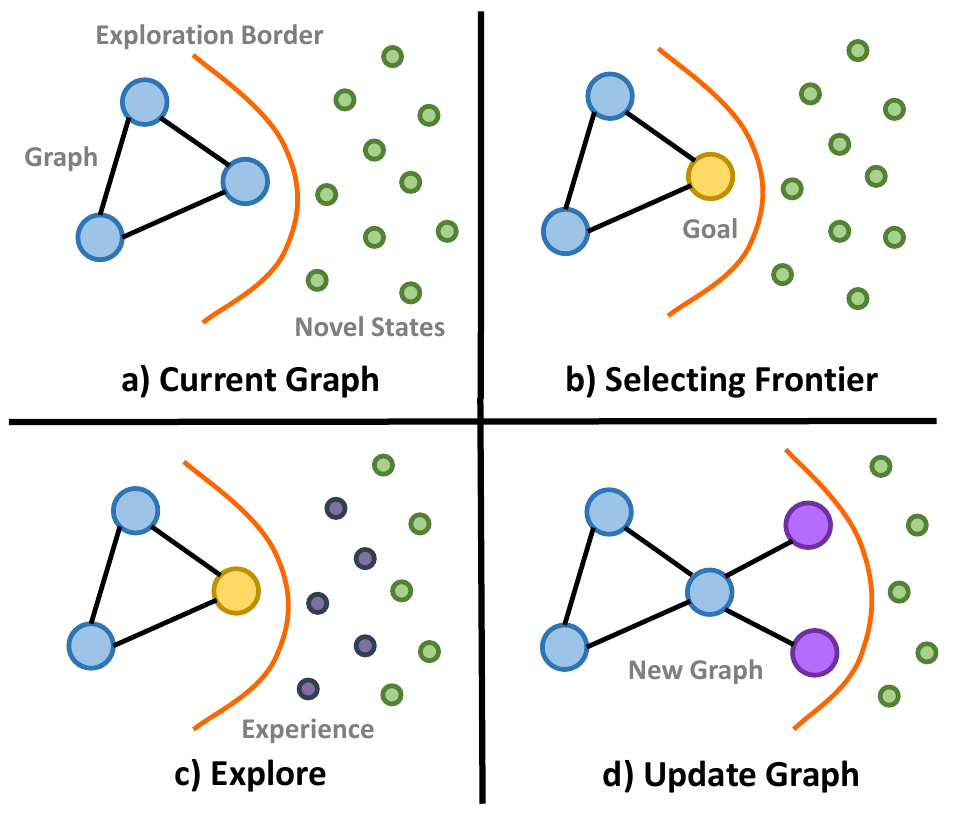}
  \caption{Exploration strategy. a) Current graph memory. b) DGMem selects a promising node as a goal and encourages navigation towards it.
           c) Novel states are more likely to be found. d) The updated version of the graph memory is obtained.}
  \label{Fig-GraphExploration}
\end{figure}

\subsection{Data-Efficient Training}
As no auxiliary feedback is available in the self-supervised navigation task, a most common reward function is a binary flag distinguishing whether the agent successfully arrives at the goal $g_{t}$.
This can be calculated by measuring the similarity between the agent observation $o_{t}$ with the the image goal $g_{t}$.
But this binary flag alone is a sparse and delayed reward signal which makes data-efficient training infeasible.
To speed up the training process, we utilize the DGMem from three aspects:
First, besides the binary success flag, we introduce two additional reward signals which are topological distance reward $r^{d}$ and navigation trajectory reward $r^{n}$.
Given an image-goal $g_{i}$, the agent's current observation $o_{t}$ and the agent's past observation $o_{t-1}$, we localize them in the DGMem graph and denote the retrieved nodes as $v^{g}_{t}$, $v^{o}_{t}$ and $v^{o}_{t-1}$.
We plan two topological path on the DGMem starting from $v^{o}_{t}$ and $v^{o}_{t-1}$ to $v^{g}_{t}$ respectively.
The difference of the path length indicates whether the robot is getting closer to the goal or not. 
Therefore, the topological distance reward $r^{d}$ is defined as follows:
\begin{equation}
    r^{d} = \alpha (|l(v_{t-1},v_{g})| - |l(v_{t},v_{g})|)
\end{equation}
To encourage the robot to traverse more areas, the navigation trajectory reward $r^{n}$ is introduced and defined as the total number count of the passing-by different nodes along the trajectory.
\begin{equation}
    r^{e} = c\cdot \mathbb{I}(v_{i})
\end{equation}
where $c$ represents the passing-by node numbers and $\mathbb{I}$ represents whether it is the first time going through a node.
And denote the succesfully arriving target reward as $r^{s}$, the entire reward function can be written as follows:
\begin{equation}
    r_{t} = r^{d}_{t} + r^{n}_{t} + r^{s}_{t}
\end{equation}

Second, as the optimal transition navigation trajectories between pairs of nodes are stored in the DGMem, it can serve as demonstration dataset for imitation learning.
In order not to intervene with RL training process, similiar to PPG\cite{Cobbe2020PhasicPG}, during update with IL objectives, we add a KL-divergence regularization to ensure the policy won't change daramtically.
Concretely, during the imitation learning phase, we sample a batch of transition trajectory from the edge set and update the following objective $J^{IL}(\theta)$:
\begin{equation}
    J^{ce}(\theta) = \hat{\mathbb{E}_{t}}[\sum_{a_{t} \in \mathcal{A}} a_{t}\log\pi_{\theta}(\hat{a}_{t}|o_{t},o_{g})] 
\end{equation}
\begin{equation}
        J^{reg}(\theta) = \hat{\mathbb{E}_{t}}[KL[\pi_{\theta_{old}}(\cdot |o_{t},o_{g}),\pi_{\theta}(\cdot |o_{t},o_{g})]]
\end{equation}
\begin{equation}
        J^{IL}(\theta) = J^{ce}(\theta) + \beta \cdot J^{reg}(\theta)
\end{equation}

\subsection{Hierachical Navigation Policy}
During the training stage, we acquire a robust image-goal navigation policy for navigating towards the recorded images in DGMem within a local area.
In the testing stage, the agent will be asked to perform long-horizon navigation tasks, given arbitrary starting and goal images.
To that end, we incorporate the planning method to compensate for the gap between training and testing.
The entire decision framework are shown in \ref{Fig-DGM}.
The agent first locates the goal state and its own state based on the provided images in the DGMem. 
To locate the images into nodes, we use the trained value functions $V(o_{o},g)$ to measure whether two states are reachable.
Denote the localized nodes as $v^{g}$ and $v^{o}_{0}$. 
Then, we use the BFS to search a shortest path leading from $v^{o}_{0}$ to $v^{g}$.
And finally, the navigation actions are executed to navigate from node to node consecutively to finish the entire long-horizon task.

\begin{figure*}[h!]
        \centering
        \includegraphics[width=1.0\textwidth]{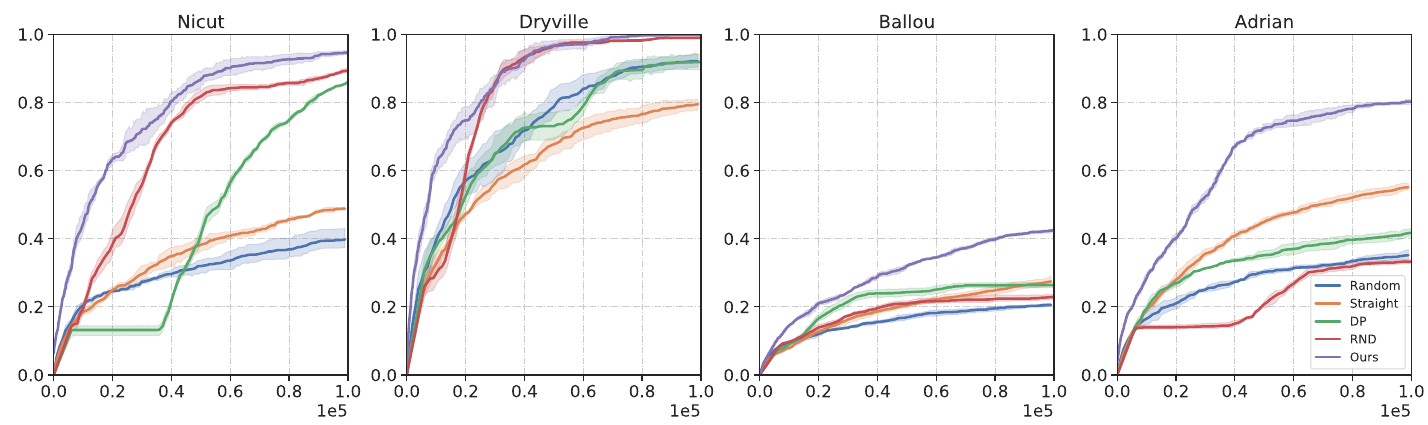}
        \caption{The coverage metric during the training stage. Only intrinsic rewards are allowed for training here.
        Our method outperforms all baselines by a large margin, especially in complicated scenarios.}
        \label{Navigation-Exploration}
\end{figure*}

\section{Experiment}
\noindent \textbf{Environment.} Our work is based on the Habitat\cite{Savva2019HabitatAP} simulator and the publicly available dataset Gibson\cite{Shen2020iGibson1A}.
We evaluate our approach on the task of image-goal navigation in eight scenes of the Habitat simulator \cite{Savva2019HabitatAP}.
The scenes vary in appearance, size, and floors. 

\noindent \textbf{Evaluation Metrics.}
We evaluate both exploration ability and navigation ability in the experiments.
The reported metrics are:
\begin{itemize}
    \item[1)] \textit{coverage}, we digitize the 3D scenes into voxels, coverage represents the proportion of traversed voxels.
    \item[2)] \textit{success rate (SR)}, the fraction of successful episodes, i.e., episodes in which agent arrives at the location with pose distance less than a threshold.
    \item[3)] \textit{success weighted by path length (SPL)}, the standard metric that weighs success by their adherence to the shortest path.
    \item[4)] \textit{distance to goal (DTS)}, the average distance to the goal pose when episode terminates.
\end{itemize}

\noindent
\textbf{Baselines.}
We compare our agent with the following four exploration methods.
\begin{itemize}
  \item \textit{Random}: An agent uses pure random policy.
  \item \textit{Straight}: An agent always select \textit{MoveAhead} but perform random rotation if collision happens.
  \item \textit{DP}: An agent uses forward dynamic prediction \cite{Pathak2017CuriosityDrivenEB} as intrinsic reward for exploration.
  \item \textit{RND}: An agent uses random network distillation \cite{Burda2018ExplorationBR} to form the intrinsic rewards.
\end{itemize}
And we compare our agent with the following methods for the evaluation of navigation policy.
\begin{itemize}
  \item \textit{BClone}: An agent uses 500 demonstration trajectories each scene for imitation learning.
  \item \textit{DAggre}: An agent that learns the optimal policy by supervised learning with an expert path planner in the simulator.
  \item \textit{PPO:} An PPO\cite{Schulman2017ProximalPO} agent uses the dense reward provided with the simulator.
  \item \textit{Sparse:} Similar to the former agent, but trained only with the sparse terminate reward. 
\end{itemize}

\begin{figure}[h!]
    \includegraphics[width=0.48\textwidth]{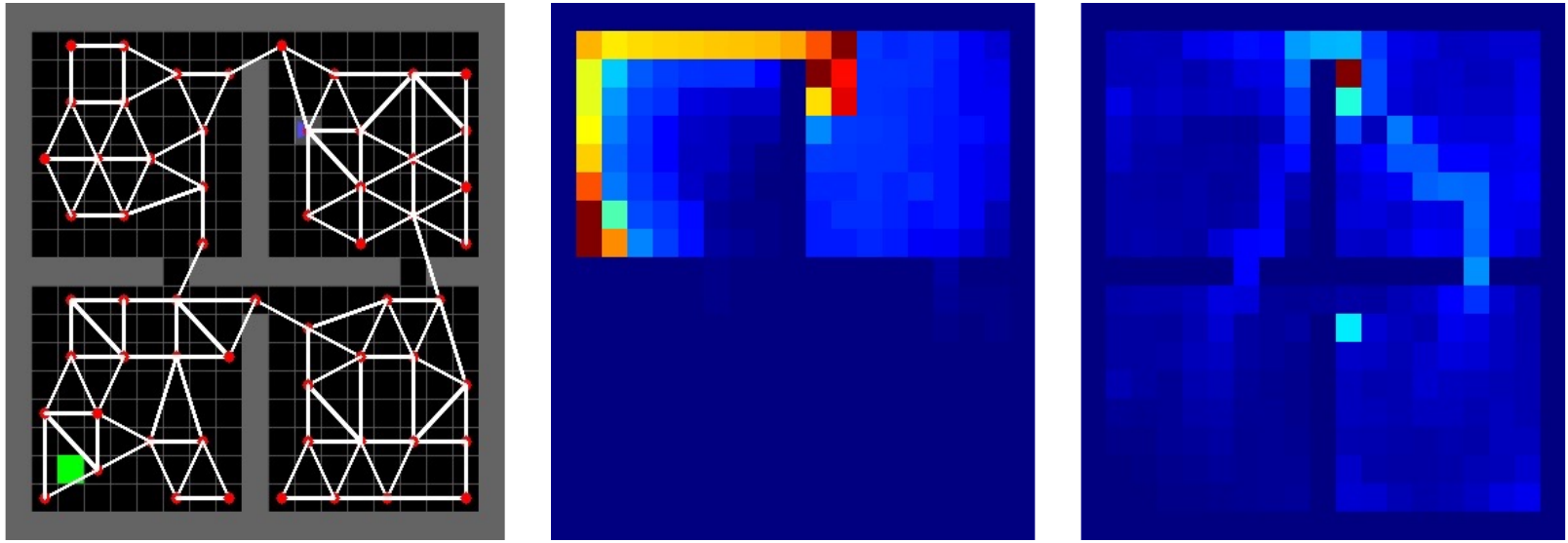}
    \caption{Visualization of the DGM results and the state distribution among all states. The middle shows the state distribution of RND method and the right shows ours.
    We limit the episode time and find it difficult for RND to discover all four rooms but roaming between two rooms which ours keeps a near-uniform distribution across all states. }
    \label{GridWorld-Exploration}
\end{figure}

\noindent
\textbf{Didactic Example.}
Before introducing our method on the visual navigation tasks, 
we first provide an intuition example on GridWorld environment.
We use FourRooms environment which contains 256 possible discrete state and 4 possible actions. 
An perfect exploration agent should try to travel across four rooms which lead to an uniform distribution among all 256 states.
We compare our DGM exploration method with RND in this example and visualize the state distribution as well as the graph memory in Fig \ref{GridWorld-Exploration}.
We limit the episode time and fix the spawn grid, so hesitation behavior (e.g move up and down) may lead to the failure arriving at all four rooms.
As shown in Fig \ref{GridWorld-Exploration}, our methods shows consistent exploration behavior and almost achieve uniform state distribution.
We also report the navigation performance in Tab~\ref{Tab:GridWorld Performance}. As the DGMem construct an uniformly distributed graph of the entire scene, the navigation policy can generalize to almost everywhere in the 
gridworld scenarios. With the goal-conditional policy only learns in a subset with 60 states, the policy generalize to all 256 states and achieves high spl (90+\%).

\begin{table}[ht]
    \centering
    \caption{Navigation Performance on GridWorld FourRooms}
    \begin{tabular}{c|c|c|c}
    \toprule[1.0pt]
     & Training Node & Success & SPL \\
    \midrule[0.5pt]
    Scene 1 & 59 & 98.1 & 95.7 \\
    \midrule[0.5pt]
    Scene 2 & 64 & 97.3 & 94.6\\
    \bottomrule[1.0pt]
    \end{tabular}
    \label{Tab:GridWorld Performance}
 \end{table}

\begin{table*}
        \centering
        \caption{Performance on indoor navigation tasks.}
        \begin{tabular}{c|c|c|c|c|c|c|c|c|c|c|c|c}
            \toprule[1pt]
            \multirow{2}{*}{Scene} & \multicolumn{3}{c}{Nicut} & \multicolumn{3}{c}{Dryville} & \multicolumn{3}{c}{Placida} & \multicolumn{3}{c}{Roane} \\
            \cmidrule(lr){2-13}
             & Success & SPL & Distance & Success & SPL & Distance & Success & SPL & Distance & Success & SPL & Distance \\
             \cmidrule(lr){1-13}
            BClone & 0.25 &	0.41 & 3.85 & 0.26 & 0.39 & 4.01 & 0.23	& 0.46 & 4.16 & 0.19 & 0.37 & 3.25 \\
            \cmidrule(lr){1-13}
            DAggre & 0.67 & \textbf{0.75} & 1.54 & 0.54 & \textbf{0.72} & 1.61 & 0.54 & \textbf{0.73} & 1.98 & 0.44 & \textbf{0.58} & 1.8 \\
            \cmidrule(lr){1-13}
            PPO & 0.07 & 0.37 & 3.77 & 0.11 & 0.43 & 3.34 & 0.25	& 0.61 & 2.62 &	0.12 & 0.33 & 2.93 \\
            \cmidrule(lr){1-13}
            Sparse & 0.01 & 0.02 & 5.14 & 0.0 & - & - & 0.0 & - & - & 0.02 & 0.03 & 4.98 \\
            \cmidrule(lr){1-13}
            Ours & \textbf{0.75} & 0.72 & \textbf{1.12} & \textbf{0.73} & 0.69 & \textbf{0.98} & \textbf{0.64} & 0.71 & \textbf{1.5} & \textbf{0.63} & 0.56 & \textbf{1.66} \\
            \bottomrule[0.5pt]
        \end{tabular}
        \vspace{2.5cm}
        \begin{tabular}{c|c|c|c|c|c|c|c|c|c|c|c|c}
            \toprule[0.5pt]
            \multirow{2}{*}{Scene} & \multicolumn{3}{c}{Adrian} & \multicolumn{3}{c}{Ballou} & \multicolumn{3}{c}{Mesic} & \multicolumn{3}{c}{Nemacolin} \\
            \cmidrule(lr){2-13}
             & Success & SPL & Distance & Success & SPL & Distance & Success & SPL & Distance & Success & SPL & Distance \\
             \cmidrule(lr){1-13}
            BClone & 0.35 & 0.47 & 2.91 & 0.04 & 0.25 & 7.06 & 0.13 & 0.36 & 5.42 & 0.08 & 0.3 & 6.52 \\
            \cmidrule(lr){1-13}
            DAggre & 0.37 & 0.58 & 2.01 & 0.07 & \textbf{0.32} & 6.19 & 0.12 & \textbf{0.47} & 4.25 & 0.05 & 0.31 & 5.96 \\
            \cmidrule(lr){1-13}
            PPO & 0.02 & 0.05 & 5.96 & 0.04 & 0.1 &	9.1 & 0.15 & 0.21 & 5.99 & 0.04	& 0.08 & 8.81 \\
            \cmidrule(lr){1-13}
            Sparse & 0.0 & - & - & 0.0 & - & - & 0.0 & - & - & 0.0 & - & - \\
            \cmidrule(lr){1-13}
            Ours & \textbf{0.61} & \textbf{0.59} & \textbf{1.77} & \textbf{0.32} & 0.28 & \textbf{3.36} & \textbf{0.41} & 0.39 & \textbf{2.96} & \textbf{0.54} & \textbf{0.39} & \textbf{3.22} \\
            \bottomrule[1pt]
        \end{tabular}
        \label{Navigation-Results}
        \vspace{-2.5cm}
    \end{table*}

\noindent
\textbf{Navigation Results.} 
We report the exploration efficiency among all methods in Fig \ref{Navigation-Exploration}.
Our approach can better explore an unknown environment compared with the curiosity modules. 
From the experiment, we observe that the curiosity-motivated agent tends to repeatedly roam over several regions. 
The uncertainty in the policy makes it difficult to perform long-range exploration skills,  
for example, traversing a long corridor, and walking upstairs to a different floor.
In contrast, our scheme owns an explicit search direction proposed by the global policy, 
which is beneficial for navigation scenarios.
Then, we evaluate the multi-goal navigation performance and the results are shown in Table \ref{Navigation-Results}.
BClone method learns the policy by fitting in the expert demonstration dataset, the scale of the demonstrations is the key factor.
With 500 trajectories for each scene, BClone shows worse generalization capability to novel goals.
DAggre method consistently correct itself by comparing the output actions with an underlying path planner.
Although this method can acquire a satisfied performance, especially in SPL.
But in many applications, we cannot possess an optimal controller in advance to collect unlimited labeled actions.
The PPO method is trained on each scene with 1M interactions. 
The poor result of the PPO indicates that even if the reward signal is dense, 
using RL to train a vanilla end-to-end architecture to finish the multi-goal image navigation tasks is difficult.
The performance gap between point-goal navigation \cite{Wijmans2020DDPPOLN} and the image navigation inspires us that how to learn an image representation to capture the spatial information and match the reward signal is an important issue for visual navigation.
And if we remove the heuristic reward used in PPO, the agent cannot learn the navigation skills at all.
Our self-supervised method, with only onboard sensor inputs, outperforms all baseline methods in success rate and distance to goal in the image-goal navigation task.
This proves the design of our approach. 

In many scenarios, DAggre achieves better SPL compared with our method.
This is mainly caused by our planning algorithm.
As we force the agent must walk through all the waypoints nodes one by one, the navigation routing becomes sub-optimcal.
Therefore, how to design a better global policy, not just using the static path-planning may be valuable for further improvements.
And how to incorporate such a graph memory for other tasks is one of our concerns in the future work.

\begin{figure}[h!]
    \includegraphics[width=0.5\textwidth]{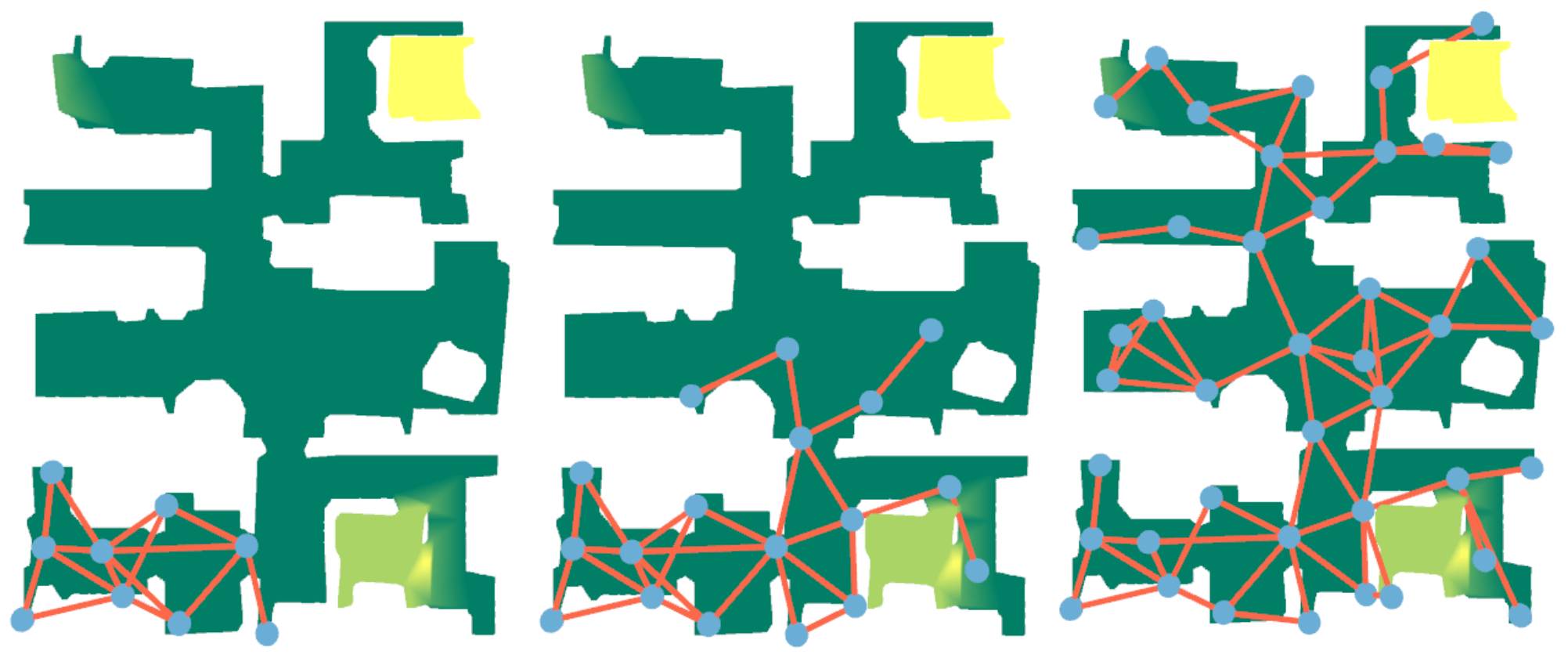}
    \caption{An example visualization of the DGM generation process. 
    At the beginning of the training process, agent roams around the spawn points, 
    so the nodes gather and some mistaken edges exists. 
    As the training goes, agents gradually travels to further areas so the graph memory grows. 
    Finally, an reasonable abstraction of the entire state spaces is generated and becomes our graph memory.}
    \label{Graph-Iteration}
\end{figure}

For better illustration of the DGM module, we give an example to show the growing process of the graph in Fig \ref{Graph-Iteration}.
As the agent interacts with the environment, the edges and nodes are dynamiclly updated and finally constructs the reasonable graph memory.

\noindent
\textbf{Ablation Study.} As we use self-localization in the training stage to build the graph memory,
we need to discuss whether our approach is robust with the pose estimation noise. 
Here, we introduce three-level scales of gaussian noise $(\tau=0.1,\tau=0.2,\tau=0.3)$ as pose estimation error and we report the average evaluation metrics across in the Table \ref{}.
The pose estimation noise only brings a minor influence on the final navigation performance.
In fact, the pose estimation only affects the quality of the graph memory. 
For example, with the pose estimation error, the robot may mistakenly adds some observations as nodes, 
and this will bring inhomogeneity to the node distribution in local areas.
And this add slighty change the training dataset and influcence the training difficulty.
But the agents can still distinguish which states are closer with the help of the value function, the value function is not related to the pose estimation error.
Therefore, the performance of the entire hierarchical policy will not degrade severely.

\begin{table}
   \centering
   \caption{Ablation Study on Pose Estimation Error}
   \begin{tabular}{c|c|c|c}
   \toprule[1.0pt]
   Noise & Success & SPL & Distance \\
   \midrule[0.5pt]
   $\tau=0.1$ & 55.0 & 52.1 & 2.29 \\
   \midrule[0.5pt]
   $\tau=0.2$ & 54.6 & 51.6 & 2.46\\
   \midrule[0.5pt]
   $\tau=0.3$ & 51.7 & 49.1 & 2.57 \\
   \bottomrule[1.0pt]
   \end{tabular}
\end{table}

\noindent
\textbf{Implementation Details} 
In our method, the visual encoder ResNet18 are pre-trained on the ImageNet dataset and we freeze the parameters during our policy training.
The fusion module concatentate the visual features from observation image and subgoal image,
and follows two fully-connected layers with size (512,256).
The GRU module contains 256 units.
We use the PPO with Adam optimizer to train the network.
we set \textit{clip ratio=0.1, nsteps=256, nminibatch=1, nepochs=4, 
learning rate=1e-4, discount factor=0.99} in our experiments.
The learning rate will linear decay to 1e-5. We train our approach with 1M steps interactions and save the best model to report the performance.
As for the hyper-parameters in our method, we set the distance threshold as 
$d_{c}=1.5$,$d_{s}=-0.85$,$d_{e}=1.0$ ,the reward coefficient as $\alpha=0.2,c=0.05,$, the imitation learning coefficient $\beta=0.1$.
We haven't carefully searched these parameters, maybe a better choice can improve the performance a little.

\section{Conclusion}
In this paper, we present the dynamic graph memory (DGMem) module for visual navigation tasks. 
Many advances in navigation tasks hinge on realistic simulators, which are costly and time-consuming to build. 
Our DGMem is one way to reuse the collected experience and form an organized memory that attempts to replace the function of simulators. 
Detailed experiments reveal that the proposed DGM enables consistent exploration and achieves better multi-goal navigation performance compared with those methods trained with supervision or heuristic rewards. 
By replacing the required attributes in simulators with DGMem, we believe our proposed method reduced the gap for real-world navigation applications.

% \subsection{}

\bibliographystyle{IEEEtran}
\bibliography{root}
\end{document}